\documentclass[pdflatex,sn-basic]{sn-jnl}

\usepackage{graphicx}%
\usepackage{multirow}%
\usepackage{amsmath,amssymb,amsfonts}%
\usepackage{amsthm}%
\usepackage{mathrsfs}%
\usepackage[title]{appendix}%
\usepackage{textcomp}%
\usepackage{manyfoot}%
\usepackage{booktabs}%
\usepackage{listings}%
\usepackage[ruled,vlined,linesnumbered]{algorithm2e}
\usepackage{algcompatible}
\usepackage{array}
\usepackage{xspace}
\usepackage{setspace}
\usepackage{hyperref}
\usepackage[dvipsnames]{xcolor}

\makeatletter
\DeclareRobustCommand\onedot{\futurelet\@let@token\@onedot}
\def\@onedot{\ifx\@let@token.\else.\null\fi\xspace}

\def\eg{\emph{e.g}\onedot} 
\def\ie{\emph{i.e}\onedot}

\newcolumntype{P}[1]{>{\centering\arraybackslash}p{#1}}
\newcolumntype{M}[1]{>{\centering\arraybackslash}m{#1}}

\begin{document}

\title[Article Title]{Towards Few-Call Model Stealing via Active Self-Paced Knowledge Distillation and Diffusion-Based Image Generation}


\author[1]{\fnm{Vlad} \sur{Hondru}}\email{vlad.hondru25@gmail.com}

\author*[1]{\fnm{Radu Tudor} \sur{Ionescu}}\email{raducu.ionescu@gmail.com}

\affil*[1]{\orgdiv{Department of Computer Science}, \orgname{University of Bucharest}, \orgaddress{\street{14 Academiei}, \city{Bucharest}, \postcode{010014}, \country{Romania}}}


\abstract{Diffusion models showcase strong capabilities in image synthesis, being used in many computer vision tasks with great success. To this end, we propose to explore a new use case, namely to copy black-box classification models without having access to the original training data, the architecture, and the weights of the model, \ie~the model is only exposed through an inference API. More specifically, we can only observe the (soft or hard) labels for some image samples passed as input to the model. Furthermore, we consider an additional constraint limiting the number of model calls, mostly focusing our research on few-call model stealing. In order to solve the model extraction task given the applied restrictions, we propose the following framework. As training data, we create a synthetic data set (called proxy data set) by leveraging the ability of diffusion models to generate realistic and diverse images. Given a maximum number of allowed API calls, we pass the respective number of samples through the black-box model to collect labels. Finally, we distill the knowledge of the black-box teacher (attacked model) into a student model (copy of the attacked model), harnessing both labeled and unlabeled data generated by the diffusion model. We employ a novel active self-paced learning framework to make the most of the proxy data during distillation. Our empirical results on three data sets confirm the superiority of our framework over four state-of-the-art methods in the few-call model extraction scenario. We release our code for free non-commercial use at \url{https://github.com/vladhondru25/model-stealing}.}




\maketitle

\section{Introduction}
\label{sec:intro}

Image classification is one of the most studied topics in computer vision. The task has been extensively investigated \citep{krizhesvky-NeurIPS-2012, he-cvpr-2016, dosovitskiy-arxiv-2020}, and as a result, there is a vast amount of open-source models that can easily be accessed, even by non-technical people. However, these are usually trained on popular data sets (\eg CIFAR-10 \citep{Krizhevsky-TECHREP-2009} or ImageNet \citep{Russakovsky-IJCV-2015}), being constrained to only predict specific object classes. If other classes are of interest, one might use a subscription-based model made available by some company, usually via a paid API. Another solution would be to train a task-specific model instead of using an already available one, the drawback being the need for a large quantity of annotated data and hardware resources.

\begin{figure}[!t]
    \centering
    \includegraphics[width=0.7\linewidth]{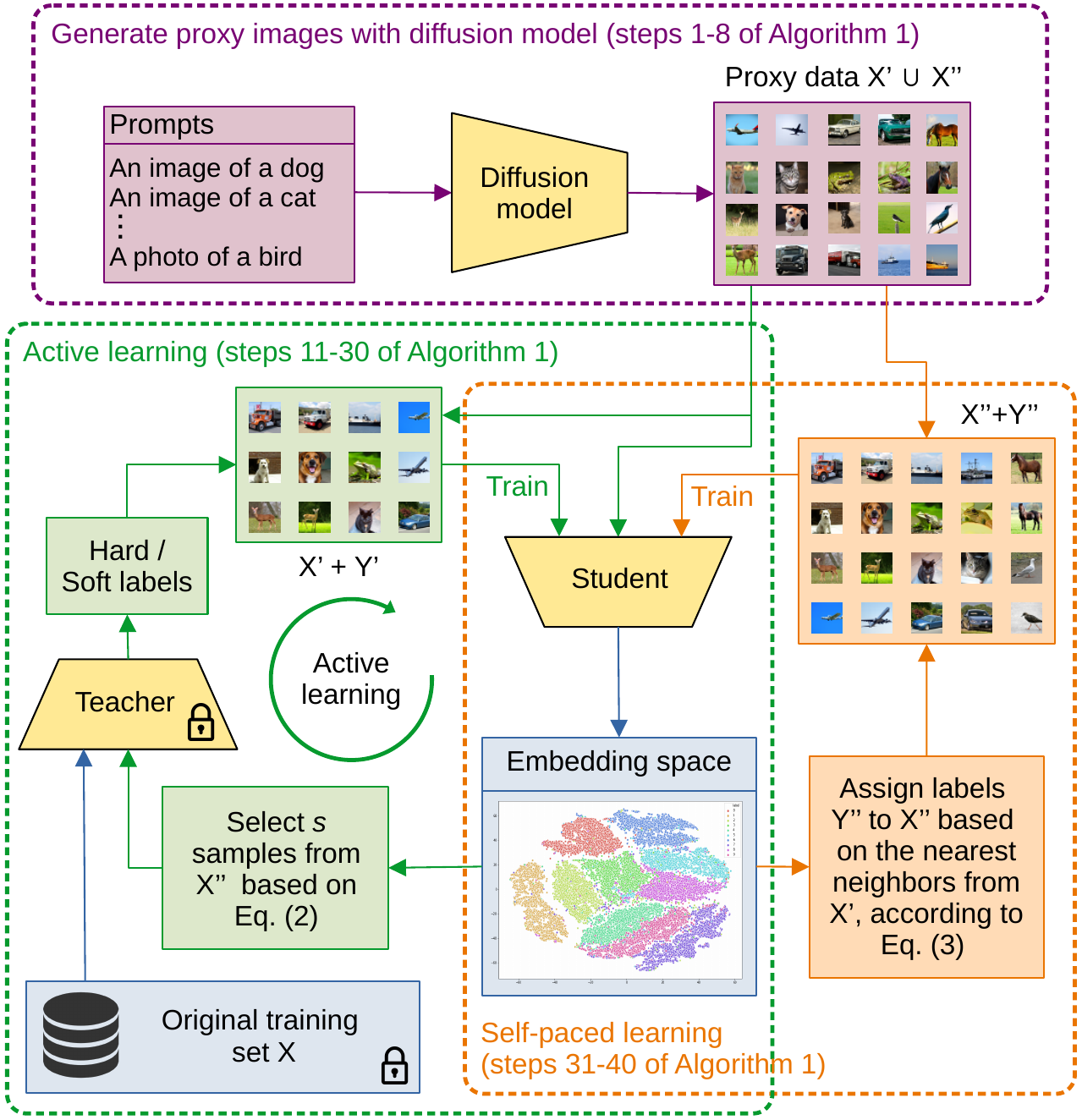}
    \caption{The proposed pipeline for model stealing starts by generating proxy images using a diffusion model. Then, proxy images are gradually annotated by the black-box teacher model and further used to train the student model via an active learning scheme. At the same time, the remaining proxy images are pseudo-labeled via a nearest neighbor scheme that operates in the latent space of the student. The pseudo-labeled images are also used to optimize the student via a self-paced learning scheme. Best viewed in color.}
    \label{fig_pipeline}
\end{figure}

With the recent AI hype, more and more individuals and businesses are eager to try or even implement AI-based solutions. In this context, enterprises ranging from small companies to large corporations have deployed deep learning models and made them publicly available. In most cases, such models are accessible as Machine Learning as a Service (MLaaS), although, on a few occasions, the models and the weights are open-sourced.  The most popular example is ChatGPT \citep{openai-chatgpt-2022}, which was made available by OpenAI. As far as the cost is concerned, a subscription-based payment scheme is often employed. Nevertheless, the facile access to the APIs results in many potential risks. One potential vulnerability is that the model's functionality can be copied \citep{tramer-usenix-2016, papernot-acm-2017, correia-IJCNN-2018, wang-SP-2018, orekondy-cvpr-2019, oh-xai-2019, barbalau-NeurIPS-2020, sanyal-cvpr-2022}, which infringes the intellectual property of the owners. As a result, exploring different methods on how to replicate black-box models will increase awareness on the existing risks, and will foster the inclusion of prevention mechanisms in the incipient development phases to counter model extraction attacks.

Given the aforementioned context, we present a pipeline that extracts the functionality of a black-box classification model (named teacher) into a locally created copy model (called student) via knowledge distillation \citep{nayak-icml-2019, micaelli-neruips-2019, yin-cvpr-2020, addepalli-AAAI-2020, barbalau-NeurIPS-2020, Chen-arxiv-2024, song-TIP-2022, Wei-ACL-2021} and self-paced active learning, as shown in Figure \ref{fig_pipeline}. Our method is deemed to be applicable in challenging real-world scenarios, where information about the data set and the training methodology (architecture, optimizer, hyperparameters, weights or other involved techniques) used to train the attacked model is completely concealed from the attacker. More precisely, our method is only able to observe the soft or hard class labels for a limited number of given input samples. 

As illustrated in Figure \ref{fig_pipeline}, the first stage of our pipeline is to employ a diffusion model to generate a proxy data set with the samples that belong to the classes of interest. 
Diffusion models \citep{croitoru-arxiv-2022} are a type of probabilistic generative models that gained a lot of traction given their ability to outperform generative adversarial networks (GANs) \citep{dhariwal-NeurIPS-2021}. These models were successfully applied to a wide variety of tasks \citep{croitoru-arxiv-2022}, ranging from unconditional image generation \citep{sohl-icml-2015, ho-NeurIPS-2020, nichol-ICML-2021a, song-ICLR-2021b}, inpainting \citep{lugmayr-CVPR-2022, Nichol-ICML-2021b} and text-to-image generation \citep{Rombach-CVPR-2022, saharia-arXiv-2022, avrahami-CVPR-2022} to image segmentation \citep{amit-arXiv-2021, baranchuk-arXiv-2021} and medical imaging \citep{wolleb-arXiv-2022a}. The wide adoption of diffusion models is determined by their capability of generating realistic, qualitative and diverse images. To the best of our knowledge, we are the first to employ diffusion models to generate proxy data for model stealing attacks. 

The next stage is to gather soft labels from the teacher model for a subset of the generated images. The size of this subset is constrained by the limited number of allowed API calls. We also consider the scenario when the black-box model returns only hard labels, showing that our pipeline is robust to the type of available labels. We propose a clustering-based approach to determine which samples are more relevant to be passed through the teacher. This is achieved by creating clusters in the latent space of the student model for each class, and computing a sampling probability for every data point, according to the distance to its corresponding cluster centroid. Then, we train our student model in a supervised setting on the labeled samples, until convergence. Finally, we introduce a self-paced learning method in which we assign pseudo-labels to the samples that were not inferred by the API due to the limited number of calls. We further train the student on a joint data set containing data samples with labels from the teacher, as well as pseudo-labeled examples. To the best of our knowledge, we are the first to study few-call model stealing.

We conduct experiments on three image data sets, CIFAR-10 \citep{Krizhevsky-TECHREP-2009}, Food-101 \citep{Bossard-ECCV-2014} and FER+ \citep{Barsoum-ICMI-2016}, considering various convolutional and transformer-based architectures for the teacher and student models. As a result of our experiments, we conclude that our pipeline generally outperforms competing methods \citep{barbalau-NeurIPS-2020,orekondy-cvpr-2019,wang-ICML-2021,zhang-ICLR-2022} by significant margins, regardless of the number of API calls. We further confirm the applicability of our method in real scenarios by showing similar efficiency, irrespective of the architecture or the type of output given by the black-box model. 

We summarize our contributions on replicating black-box classification models below:
\begin{itemize}
    \item We harness diffusion models to create synthetic proxy data sets consisting of relevant samples for model stealing attacks.
    \item We propose a novel strategy on how to actively choose the samples for which to collect labels from the attacked model, obtaining improved results in the few-call model stealing scenario. 
    \item We introduce a novel strategy that assigns pseudo-labels to the remaining samples and uses them to further boost the performance of the student via self-paced learning.
\end{itemize}

\section{Related Work}
\label{sec:related_work}

The model stealing research directions can be divided into different categories from multiple perspectives. For instance, related studies can be divided based on their main goal into attacking methods \citep{barbalau-NeurIPS-2020,correia-IJCNN-2018,orekondy-cvpr-2019,papernot-acm-2017,sanyal-cvpr-2022,wang-ICML-2021,zhang-ICLR-2022} and defense methods \citep{juuti-essp-2019,kesarwani-csac-2018,liu-esrcs-2022,Haonan-IEEE-2022,ye-AAAI-2022,zhang-acm-2021}. Another organization is given by the trade-off between accuracy \citep{barbalau-NeurIPS-2020,correia-IJCNN-2018,shi-hst-2017,wang-ICML-2021} and number of API calls \citep{chandrasekaran-usenix-2020,li-ICDM-2018,shi-hst-2018,tramer-usenix-2016,zhang-ICLR-2022}. Moreover, some studies \citep{wang-SP-2018, oh-xai-2019, tramer-usenix-2016} are aimed at retrieving exact information about the attacked model, \eg~its architecture or its hyperparameters, while others \citep{shi-hst-2017, jagielski-arXiv-2019, orekondy-cvpr-2019, barbalau-NeurIPS-2020} are aimed at mimicking its behavior. Two important categories in which the model stealing methods can be divided are given by the data used for training. Some methods \citep{papernot-acm-2017, correia-IJCNN-2018, pal-arxiv-2019} use real data, just as the attacked model, while others \citep{barbalau-NeurIPS-2020,kariyappa-cvpr-2021,mosafi-IJCNN-2019,orekondy-cvpr-2019,sanyal-cvpr-2022,wang-ICML-2021,xie-esrcs-2022,zhang-ICLR-2022} assume the training data is not accessible, resorting to artificially generating proxy data. We refer the readers to the survey of \cite{oliynyk-acm-2023}, who presented a comprehensive taxonomy comprising multiple model stealing methods, which are described in great detail. We next concentrate on closely related studies that are replicating black-box models by launching attacks, while taking into consideration the balance between accuracy and number of model queries.

In one of the earliest works in this area, \cite{tramer-usenix-2016} showed how to extract the capability of a black-box model, but instead of just copying the functionality, their goal was to approximate the parameters. In order to carry out such a strict task, they assumed some prior insight about the model type and training data. They advocate using the exact classes (hard labels) as the output of the attacked models to greatly improve the prevention of stealing attacks. In contrast, we demonstrate similar performance levels irrespective of the output type (soft or hard), while preserving the black-box nature of the teacher. 

A stepping stone in model stealing research was the paper from \cite{papernot-acm-2017}, which presented a method that had a similar setting as ours, but with a different objective: instead of trying to fully replicate the black-box model functionality with high accuracy, they are approximating the decision boundary. While also leveraging synthetic data obtained by augmenting some part of the original data set (thus weakening one of our assumptions), their aim is to only launch adversarial attacks. Related efforts have been made by \cite{biggio-springer-2013} and \cite{goodfellow-arxiv-2014}, but with even weaker constraints on knowledge about the data and the teacher model. 

\cite{barbalau-NeurIPS-2020} developed a framework, called Black-Box Ripper, that generates samples using GANs and then optimizes the samples with an evolutionary algorithm until the images become relevant, \ie~produce a high response from the teacher model. Although the presented results showed better performance than alternative approaches, the authors assumed a relaxed setting, in which an unbounded number of API calls is permitted. 
For a fair comparison with Black-Box Ripper, we consider the same number of API calls for both Black-Box Ripper and our framework.

Similar to Black-Box Ripper \citep{barbalau-NeurIPS-2020}, \cite{sanyal-cvpr-2022} and \cite{xie-esrcs-2022} leveraged GANs to create synthetic samples which are subsequently used in launching stealing attacks. Nevertheless, the latter authors simultaneously trained the generative and the discriminative models, thus continuously improving the quality of the artificial data. \cite{sanyal-cvpr-2022} demonstrated their method for a larger number of classes (100), as well as using only hard labels. \cite{xie-esrcs-2022} chose a different approach by implementing an active learning strategy for the classes to be sampled by the GANs.

With the same objective as our work, \cite{orekondy-cvpr-2019} introduced Knockoff Nets, an approach to replicate a deep learning model made available as MLaaS, focusing at the same time on being mindful with respect to the number of queries. They utilized a large-scale proxy data set, namely ImageNet \citep{Russakovsky-IJCV-2015}, and, in order to make as few API calls as possible, they employed a reinforcement learning strategy that trains a policy to choose the more relevant samples.

\cite{wang-ICML-2021} introduced a method to distill the knowledge from a black-box model in similar conditions to ours, although their call budget is much larger (they use between 25 thousand and 2 million calls). As the training data is not available, synthetic data is being iteratively generated. Starting from random noise, the hard labels are obtained by querying the teacher model and then optimizing the samples. Finally, soft pseudo-labels are computed using various methods that measure sample robustness, and then, the student model is trained with the respective targets.

\cite{yin-cvpr-2020}, \cite{Fang-IJCAI-2021} and \cite{Li-IS-2023} studied data-free knowledge distillation. \cite{yin-cvpr-2020} proposed DeepInversion, a method to generate images from the distribution used to train a teacher model. The method relies on the information stored in the batch normalization layers of the pre-trained teacher. \cite{Fang-IJCAI-2021} focused on generating a diverse set of samples for a pre-trained teacher model, based on the conjecture that higher data diversity should correspond to stronger instance discrimination. \cite{Li-IS-2023} adopted curriculum learning \citep{Soviany-IJCV-2022} to obtain a dynamic training strategy that gradually adjusts the complexity of the generated data samples. While the assumption of data-free knowledge distillation frameworks is that the original training data is not known, such frameworks have full access to the teacher model. Our approach is designed for a more restrictive setup, where the architecture of the pre-trained teacher is not known at all, which prevents gradient propagation through the teacher. In summary, data-free knowledge distillation frameworks are not directly applicable to the black-box model stealing task, since they operate in a more relaxed setup.

Recently, \cite{zhang-ICLR-2022} employed a generator to create synthetic samples. The generator is trained with a cross-entropy loss (to obtain qualitative samples) and an information entropy loss (to obtain diverse samples). Its weights are randomly reinitialized at the beginning of each epoch, while the best generated samples are kept until the end. The labels of the saved samples are obtained from the black-box teacher model, and then, the student is trained with these. 

Different from the aforementioned related works, we do not require any additional information to obtain the proxy data samples. Moreover, we take a step further in regards to the number of permitted API calls, and not only try to minimize them, but rather have a fixed low number of queries. 
To the best of our knowledge, we are the first to propose a few-call model stealing framework that is applicable in all respects to a real model theft scenario.

\section{Method}
\label{sec:method}

We begin by presenting the studied task and continue by introducing our method for replicating black-box models, while describing our novel components and how to integrate them in the proposed framework.

\subsection{Problem Statement} 
As stated in previous works \citep{orekondy-cvpr-2019}, the model stealing task is very similar to knowledge distillation, \ie in both cases, the task is to infuse the functionality of a teacher model into a student model. However, the goal of knowledge distillation is to produce a compressed model with comparable accuracy, which is different from the goal of black-box model stealing. In the context of model stealing, we assume no knowledge about the training data, the architecture and the weights of the teacher. Moreover, the student architecture is not required to be less complex. Nevertheless, in a similar manner, we refer to the black-box model as the teacher, and the copy model as the student. 

Black-box models are usually available as MLaaS. In a real scenario, service providers do not disclose any information about the model. The training data, the architecture of the model, its weights, gradients or hyperparameters, and other related details are unknown to the MLaaS users. Furthermore, for each query, the providers only supply the output of the model, either as soft labels (class probabilities) or  hard labels. We consider an even more strict scenario where the number of queries is limited due to the following consideration: the model stealing attack might get detected due to the high number of API calls. Moreover, even if the attack remains undetected, the costs might rise to unjustifiable levels after a certain number of API calls.

Formally, we can formulate the problem statement using the following objective: 
\begin{equation}
  \min_{\theta_S} \lVert T(X, \theta_T) - S(X' \cup X'', \theta_S) \rVert,
  \text{ subject to } \lvert X'\rvert \leq n,
\label{eq:formal_problem}
\end{equation}
where $T$ is the black-box teacher model, $S$ is the student model in which we distill the knowledge, $\theta_T$ and $\theta_S$ are their corresponding weights, $X$ is the original data set, while $X'$ and $X''$ represent the two parts for the synthetic (proxy) data set, namely the part labeled by $T$ and the part with pseudo-labels. The aim is to optimize the student weights such that the difference between the outputs of the two models is negligible, subject to making at most $n$ passes through the teacher $T$, \ie $n$ represents the number of API  calls. Following previous work \citep{addepalli-AAAI-2020,barbalau-NeurIPS-2020,orekondy-cvpr-2019, zhang-ICLR-2022}, instead of using models accessible via APIs, we train the teacher ourselves, prior to launching the attack. During the attack, we use the teacher in a black-box regime, thus preserving all the constraints mentioned above. We hereby attest that no information about the teacher is leaked while training the student.

\begin{algorithm*}[!th]
\scriptsize{
\caption{Active Self-Paced Knowledge Distillation (ASPKD)}
\label{alg:method}
\begin{spacing}{0.85}
\KwIn{$T$ - black-box model, $S$ - student model, $G$ - diffusion model, $m$ - number of proxy samples, $\mathcal{C}$ - set of classes, $n$ - maximum number of teacher calls, $s$ - number of calls per iteration ($s \leq n$), $k$ - number of neighbors.}
\KwOut{
$\theta_S$ - trained weights of the student (copy) model.
}
$\mathcal{T} \leftarrow \{\text{``An image of a \%s'', An photo of a \%s''}\};$ $\lhd$ initialize the set of templates

$X' \leftarrow \emptyset, Y' \leftarrow \emptyset;$ $\lhd$ initialize first proxy training subset 

$X'' \leftarrow \emptyset, Y'' \leftarrow \emptyset;$ $\lhd$ initialize the second proxy training subset 

\ForEach{$i \in \{1,2,...,m\}$}
{
    $c \sim \mathcal{U}(\mathcal{C});$ $\lhd$ randomly sample a class label from a uniform distribution 
    
    $t \sim \mathcal{U}(\mathcal{T});$ $\lhd$ randomly sample a prompt template
    
    $x''_i \leftarrow G(t \text{ \% str}(c));$ $\lhd$ generate an image for some text prompt
    
    $X'' \leftarrow X'' \cup \{ x''_i \}, Y'' \leftarrow Y'' \cup \{ c \};$ $\lhd$ add generated image and label 
    
}
$\theta_S \sim \mathcal{N}\left(0,{2}/({d_{in} + d_{out}})\right);$ $\lhd$ initialize weights of student using Xavier initialization

\Repeat{$\lvert X' \rvert = n$}{
    $Z'' \leftarrow \emptyset;$ $\lhd$ initialize the set of latent vectors
    
    $\mu_c \leftarrow \mathbf{0}_d, \nu_c \leftarrow 0, \forall c \in \mathcal{C};$ $\lhd$ initialize  centroids and number of samples per class 
    
    \ForEach{$i \in \{1,2,...,\lvert X''\rvert\}$}{
        $z''_i \leftarrow \bar{S}(x''_i, \theta_S));$ $\lhd$ obtain the latent vector for sample $x''$
        
        $Z'' \leftarrow Z'' \cup \{ z''_i \};$ $\lhd$ add latent vector to the set $Z''$
        
        $\mu_{y''_i} \leftarrow \mu_{y''_i} + z''_i, \nu_{y''_i} \leftarrow \nu_{y''_i} + 1;$ $\lhd$ add latent vector to centroid of class $y''_i$ 
    }
    $\mu_c \leftarrow \frac{\mu_c}{\nu_c}, \forall c \in \mathcal{C};$ $\lhd$ compute the centroids for all classes
    
    $\mathcal{P} \leftarrow \emptyset;$ $\lhd$ initialize the set of probabilities for inclusion in $X'$
    
    \ForEach{$i \in \{1,2,...,\lvert X''\rvert\}$}{
        $c \sim \mathcal{U}(\mathcal{C});$ $\lhd$ randomly sample a class label from a uniform distribution 
        
        $p_i \leftarrow \exp\left(-{{\Delta\left(\bar{S}(x''_i), \mu_c\right)}/({2\cdot \sigma^2})}\right);$ $\lhd$ apply Eq.~\eqref{eq:dist2prob}
        
        $\mathcal{P} \leftarrow \mathcal{P} \cup \{ p_i \};$ $\lhd$ add probability to $\mathcal{P}$
    }
    \ForEach{$i \in \{1,2,...,\min\{s, n - \lvert X' \rvert\}\}$}{
        $x'_i \sim \mathcal{P}(X'');$ $\lhd$ sample image using the probability distribution of $\mathcal{P}$
        
        $y'_i \leftarrow T(x'_i);$ $\lhd$ obtain the target label from the teacher
        
        $X' \leftarrow X' \cup \{ x'_i \}, Y' \leftarrow Y' \cup \{ y'_i \};$ $\lhd$ add image and teacher label 
        
        $X'' \leftarrow X'' - \{ x'_i \}, Y'' \leftarrow Y'' - \{ y'_i \};$ $\lhd$ remove image and label 
        
    }
    \Repeat{convergence}{
        \ForEach{$i \in \{1,2,...,\lvert X'\rvert\}$}{
            $\theta_S \leftarrow \theta_S - \eta \cdot \nabla \mathcal{L}(x'_i, y'_i, \theta_S);$ $\lhd$ train student on $X'$ with labels $Y'$ 
        }
    }
    \ForEach{$x''_i \in X''$}{
        $D \leftarrow \mathbf{0}_{\lvert X'\rvert};$ $\lhd$ initialize the vector of distances with zeros
        
        \ForEach{$x'_j \in X'$}{
         $d_j \leftarrow \Delta_{cos}\left(\bar{S}(x''_i), \bar{S}(x'_j) \right);$ $\lhd$ apply Eq.~\eqref{eq_cosdist} and store distance to $d_j$
        }
        $*, I \leftarrow \mbox{sort}(D);$ $\lhd$ sort the distances (in ascending order) and return indexes
        
        $y''_i \leftarrow \sum_{j=1}^k (1-d_{I_j}) \cdot y'_{I_j};$ $\lhd$ assign label to $x''_i$ based on a weighted average 
  }
  \Repeat{convergence}{
        \ForEach{$(x,y) \in (X' \cup X'', Y' \cup Y'')$}{
            $\theta_S \leftarrow \theta_S - \eta \cdot \nabla \mathcal{L}(x, y, \theta_S);$ $\lhd$ train student on $X' \cup X''$ with labels $Y' \cup Y''$ 
        }
    }
}
\end{spacing}
}
\end{algorithm*}

\subsection{Overview} 
Our framework comprises three stages, as illustrated in Figure \ref{fig_pipeline}. In the first stage, proxy images are generated by a diffusion model. In the second stage, a number of proxy images are passed to the teacher and the resulting labels are used to train the student via knowledge distillation. In the third stage, the left proxy samples are pseudo-labeled via a nearest neighbors scheme. The second and third stages are repeated in a loop until $\lvert X' \rvert = n$, thus generating a novel active self-paced knowledge distillation (ASPKD) framework. The three stages are formally integrated into Algorithm \ref{alg:method}. We next describe the individual stages, referring to the corresponding steps of the algorithm along the way.

\subsection{Data Generation}
The first challenge to overcome in order to address black-box model stealing is to procure training data. One solution is to leverage generative models to create synthetic data. While previous works \citep{barbalau-NeurIPS-2020,sanyal-cvpr-2022} used GANs~\citep{Goodfellow-NIPS-2014} to generate proxy data samples, we resort to the use of diffusion models.
Since we need to generate instances of specific object classes, we opt for text-conditional diffusion models. 
To thoroughly validate our method, we employ three different diffusion models: Stable Diffusion \citep{Rombach-CVPR-2022}, GLIDE \citep{Nichol-ICML-2021b} and SDXL \citep{podell-sdxl-2023}. Stable Diffusion and SDXL are based on a latent diffusion model, where the diffusion process is carried out in the latent space of a U-Net auto-encoder. The U-Net integrates a cross-attention mechanism to condition the image synthesis on text. GLIDE is a diffusion model that can alternate between two guidance methods, a classifier-free method and CLIP-based method \citep{radford-icml-2021}. In our approach, we select the former option. 
We use the publicly released GLIDE model, which was trained on a heavily filtered data set.

As far as the prompts are concerned, for each class, we consider two alternative prompt templates, namely ``An image of a \emph{\{class\}}'' and ``A photo of a \emph{\{class\}}'' (step 1 of Algorithm \ref{alg:method}). To generate a concrete prompt (step 7 of Algorithm \ref{alg:method}), the placeholder \emph{\{class\}} is replaced with an actual class name, \eg \emph{dog}, \emph{car}, etc. According to step 6, about half of the images of each class are generated using the first prompt, while the other half using the second prompt. This prompt variation is supposed to induce a higher variability in the generative process, thus obtaining more diverse images. However, we note that the primary factor that induces a high variability of the generated samples is the noise image that represents the starting point of the reverse diffusion process, which is randomly generated each time a new image is sampled.

\subsection{Active Learning} 
An important factor for achieving a high accuracy while having a constraint on the number of API calls is the choice of samples to pass through the black-box model. Not only should the chosen images be representative, but they should also be diverse to improve the generalization capacity of the student. To this extent, we propose an active learning methodology to select representative and diverse samples to be inferred by the teacher.

At any iteration of the active learning procedure, we obtain the latent vectors of all samples from the proxy subset $X''$, which are returned by the student (steps 14-15 of Algorithm \ref{alg:method}). Based on the labels assigned by the student, we cluster the samples into classes and compute the centroid of each class (steps 16-17 of Algorithm \ref{alg:method}). Next, we employ a sampling strategy that promotes the selection of examples closer to the centroids (steps 18-21 of Algorithm \ref{alg:method}). The idea behind our strategy is to demote the selection of outliers, since these are more likely to be mislabeled by the teacher. We exploit the distance from each latent vector to its nearest centroid in a Radial Basis Function to compute the probability of sampling the corresponding image, as follows: 
\begin{equation}
p_i = \exp\left(-{\frac{\Delta\left(\bar{S}(x''_i), \mu_c\right)}{2\cdot \sigma^2}}\right),
\label{eq:dist2prob}
\end{equation}
where $\mu_c$ is the closest centroid to $\bar{S}(x''_i)$, and $\sigma$ is a hyperparameter that controls the importance of the proximity to the nearest centroid. We select a uniformly distributed number of samples from each cluster to ensure the diversity of samples. The selected samples are given as input to the teacher, which returns a soft or hard label that is stored in $Y'$ (steps 25-26 of Algorithm \ref{alg:method}). We use the subset labeled by the teacher, denoted as $X'$, to train the student until convergence (steps 28-30 of Algorithm \ref{alg:method}). 

\subsection{Self-Paced Learning} 
Since we assume that there is a limit imposed on the number of API calls, we can only retrieve class labels from the black-box model for a fraction of our proxy data set. Hence, we have a large subset $X''$ of data samples that have not been labeled by the teacher. We propose a self-paced knowledge distillation procedure, in which we leverage the unlabeled data to improve the student. Our self-paced learning procedure gradually assigns labels to the remaining data using a nearest neighbor procedure applied in the latent embedding space learned by the student (steps 32-37 of Algorithm \ref{alg:method}). 
More precisely, we operate in the latent space of the layer right before the flattening operation or the global average pooling layer, depending on the backbone architecture of our student. Let $\bar{S}$ denote the latent space encoder.
The first step of the proposed self-paced learning method is to pass each example from the annotated proxy subset $X'$ through the student model and store the latent vectors and the labels assigned by the student. Then, for each unlabeled image $x''_i \in X''$, we gather its corresponding latent representation and search for the closest $k$ samples from the annotated training set (step 36 of Algorithm \ref{alg:method}). To compute the distance in the latent space between two samples $x''_i \in X''$ and $x'_j \in X'$, we consider two alternative metrics, namely the Euclidean distance and the cosine distance. The latter is computed as follows:
\begin{equation}\label{eq_cosdist}
\Delta_{cos}\left(\bar{S}(x''_i), \bar{S}(x'_j) \right) = 1 - \frac{\langle \bar{S}(x''_i), \bar{S}(x'_j)\rangle}{\lVert \bar{S}(x''_i)\rVert\cdot\lVert \bar{S}(x'_j)\rVert},
\end{equation}
where $\langle \cdot, \cdot\rangle$ denotes the scalar product, and $\bar{S}$ is the student encoder. Then, the label assigned to the sample $x''_i$ is inferred from the labels of its $k$ nearest neighbors (step 37 of Algorithm \ref{alg:method}). We emphasize that for the self-paced learning stage we do not use an explicit difficulty score and we do not select the samples based on difficulty. Eq.~{\eqref{eq_cosdist}} is simply used to find nearest neighbors from the labeled set $X'$. The selected neighbors further provide pseudo-labels for the samples in $X''$. These pseudo-labels are updated as the model learns, hence the self-paced learning.

For the label assignment step, we suggest two schemes, depending on the output type received from the teacher. If the teacher provides soft labels, the resulting class distribution of image $x''_i$ is computed as a weighted average of the soft labels, where the weight of a sample $x'_j$ is inversely proportional to the distance between $x''_i$ and $x'_j$. If the black-box model returns hard labels, we adopt a voting scheme based on plurality (majority) voting, where the distances between $x''_i$ and its neighbors are used to break ties. During self-paced training, some of the samples $x''_i \in X''$ are filtered out based on the following heuristic. Knowing the original class used to conditionally generate a proxy image, we eliminate the respective image if the original class is different from the assigned pseudo-label. We empirically observed that this filtering procedure generally eliminates a low number of images (around 2\%). However, by leveraging prior information from the generation step, we increase the probability of the label assignment being correct, which is especially useful in scenarios with few examples. Finally, the student model is trained on the proxy data $X' \cup X''$ (steps 38-41 of Algorithm \ref{alg:method}).

The active self-paced learning procedure is executed for a number of $r=n/s$ steps, until the API limit $n$ is reached. We note that the latent space of the student changes during training. Hence, the latent vectors are computed at every step of the active learning procedure, to ensure that the sample selection procedure is on par with the current state of the student. During our experiments, we set $r=3$, except for the one-shot and two-shot experiments, where $r$ is constrained to $r=1$ and $r=2$, respectively.


\begin{figure*}[!t]
\begin{center}
\centerline{\includegraphics[width=1.0\linewidth]{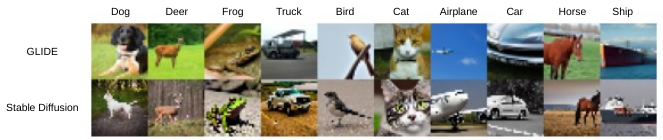}}
\caption{Samples of generated images by GLIDE \citep{Nichol-ICML-2021b} (top row) and Stable Diffusion \citep{Rombach-CVPR-2022} (bottom row) for the CIFAR-10 classes.}
\label{glide_stable_examples}
\end{center}
\end{figure*}

\begin{figure*}[!t]
\begin{center}
\centerline{\includegraphics[width=0.75\linewidth]{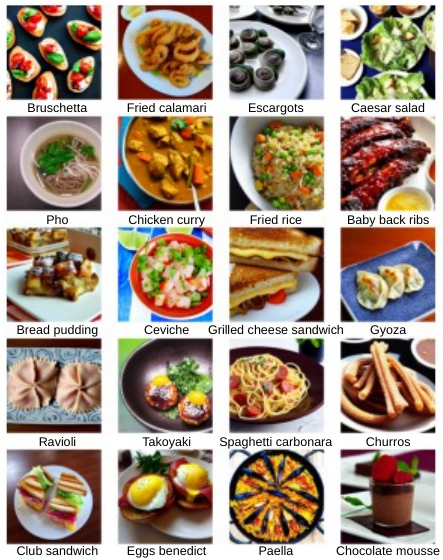}}
\caption{Samples of generated images by Stable Diffusion \citep{Rombach-CVPR-2022} for some of the Food-101 classes.}
\label{food101_examples}
\end{center}
\end{figure*}

\section{Experiments}
\label{sec:experiments}

\subsection{Experimental Setup}

\subsubsection{Data Sets}
We conduct experiments on three image data sets, namely CIFAR-10 \citep{Krizhevsky-TECHREP-2009}, Food-101 \citep{Bossard-ECCV-2014} and FER+ \citep{Barsoum-ICMI-2016}. CIFAR-10 is a data set of 50,000 training images and 10,000 test images, representing objects from 10 categories. Each image has a resolution of $32\times 32$ pixels. Food-101 \citep{Bossard-ECCV-2014} is a data set containing images of 101 food categories. Each image has a resolution of $224\times 224$ pixels. The original split contains 75,750 training images and 25,250 test images. FER+ is a curated version of the FER 2013 data set \citep{Goodfellow-ICONIP-2013}, containing images of faces showing eight different facial expressions. FER+ includes 27,473 training images and 7,092 test images, all having a resolution of $48\times 48$ pixels. These data set choices are aimed at testing the model stealing frameworks in distinct settings, comprising both low-resolution and high-resolution images, as well as a small and a large number of classes. The training sets are only used to train the black-box teachers. In contrast, the copy models are trained on generated proxy data.

\subsubsection{Diffusion Models and Proxy Data} 

In general, the proxy data should contain images with object classes for which we aim to copy the functionality of the black-box models. For instance, if we want to obtain a student that replicates the teacher on \emph{cat} and \emph{dog} classes, the proxy data should contain images of cats and dogs. For demonstration purposes, we choose to copy the full set of classes in CIFAR-10, Food-101 and FER+ data sets.

We generate two proxy data sets for CIFAR-10, one with Stable Diffusion v2 \citep{Rombach-CVPR-2022} and one with GLIDE \citep{Nichol-ICML-2021b}. The generated images are resized to match the input size of $32 \times 32$ pixels, as required by the black-box teacher. For each proxy data set, we generate 5,000 images per class. In Figure~\ref{glide_stable_examples}, we present one generated sample per class from each proxy data set.

For Food-101, we generate a proxy data set with 1000 images per class, using Stable Diffusion v2. We illustrate some randomly chosen synthetic images from this data set in Figure~\ref{food101_examples}.

\begin{figure}[!t]
\begin{center}
\centerline{\includegraphics[width=0.75\linewidth]{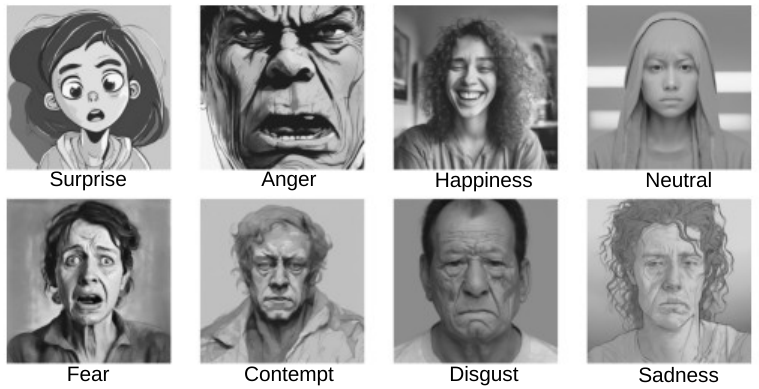}}
\caption{Samples of generated images by SDXL \citep{podell-sdxl-2023} for the FER+ classes.}
\label{ferplus_examples}
\vspace{-0.6cm}
\end{center}
\end{figure}

To generate synthetic data for FER+, we employ SDXL \citep{podell-sdxl-2023}. We generate 4,000 samples for each facial expression class. Some synthetic samples from the proxy data set are presented in Figure~\ref{ferplus_examples}.

\begin{figure}[!t]
\begin{center}
\centerline{\includegraphics[width=1.0\linewidth]{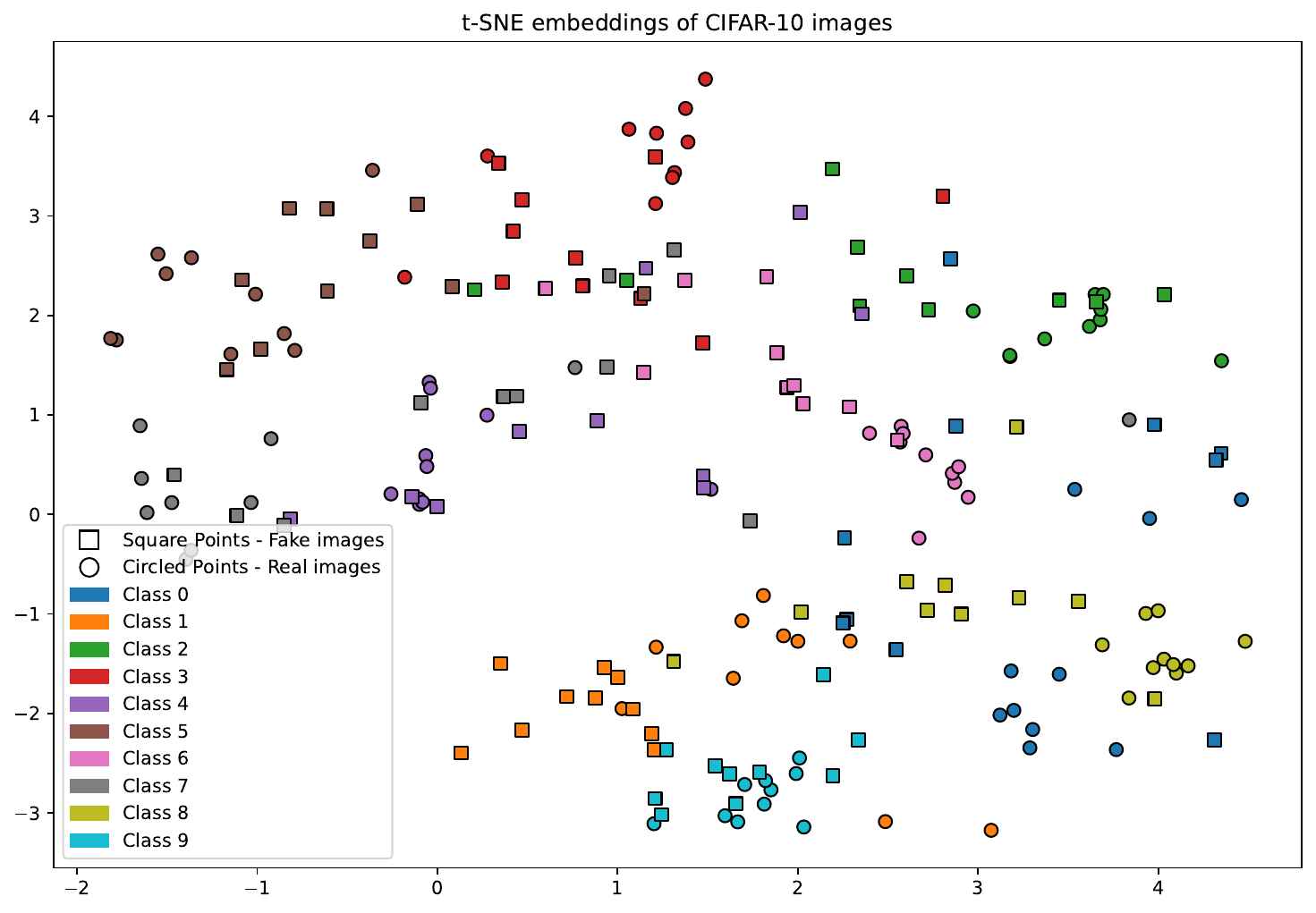}}
\caption{Real and synthetic CIFAR-10 images plotted using t-SNE. There is a high overlap between the real and proxy data.}
\label{tsne}
\end{center}
\end{figure}

Although we can easily generate many more proxy images, we choose to limit the number of generated images in each proxy data set to the number of samples available in the original CIFAR-10, Food-101 and FER+ data sets. For each proxy data set, we keep $15\%$ of the generated images for validation purposes.

For optimal results, the distribution of the proxy data should be as close as possible to that of the real data in terms of realism and diversity. We apply t-SNE, an unsupervised non-linear dimensionality reduction technique, over the latent representation of a pre-trained ResNet-50 model to compare the distributions of real samples from CIFAR-10 and proxy data generated by Stable Diffusion v2. The compared distributions are illustrated in Figure~\ref{tsne}. We observe that the proxy data distribution closely models the distribution of the real data, which is more likely to help the student model in the distillation process. To further attest the quality of the generated data, we compute the Fr\'echet Inception Distance (FID) for the proxy data, as well as the Inception Score (IS) for both real and proxy data. We report the corresponding quantitative results in Table~\ref{tab:fid_is_scores}. We observe that the quality of the synthetic data is generally high, since the diffusion models achieve IS values similar to the real data.

\begin{table*}[t]
\centering
\setlength\tabcolsep{0.18cm}
\caption{Fr\'echet Inception Distance (FID) and Inception Score (IS) computed for the real versus proxy data sets generated by diffusion models. Lower FID and higher IS values indicate better results.}
\label{tab:fid_is_scores}
\begin{tabular}{llccc}
    \toprule
    {Data set} & {Diffusion model} & 
    {FID ($\downarrow$)} & 
    {IS ($\uparrow$) on fake data} & 
    {IS ($\uparrow$) on real data} \\
    \midrule
    \multirow{2}{*}{CIFAR-10} & GLIDE & $35.75$ & $10.15\pm0.35$ & $ 10.97\pm0.29$ \\
     & SDv2 & $47.45$ & $10.34\pm0.18$ & $ 10.97\pm0.29$ \\
    \midrule
    Food-101 & SDv2 & $30.58$ & $7.87\pm0.14$ & $ 10.97\pm0.40$ \\
    \midrule
    FER+ & SDXL & $173.53$ & $3.31\pm0.06$ & $ 4.07\pm0.15$ \\
    
  \bottomrule
\end{tabular}
\end{table*}

\subsubsection{Teacher and Student Models}

In our experiments, we employ well-known model architectures that are fundamental to research, as this facilitates comparison with other baselines. For the black-box models, we use three architectures: AlexNet \citep{krizhesvky-NeurIPS-2012}, ResNet-50 \citep{he-cvpr-2016} and ViT-B \citep{dosovitskiy-arxiv-2020}. Following previous research on model stealing \citep{addepalli-AAAI-2020,barbalau-NeurIPS-2020}, we consider lighter student architectures. For the AlexNet teacher, the student is Half-AlexNet, an architecture where the number of convolutional filters and the number of neurons in fully-connected layers are reduced by $50\%$. For the ResNet-50 teacher, the corresponding student is ResNet-18 \citep{he-cvpr-2016}. Finally, for the ViT-B teacher, we select FastViT \citep{vasu-iccv-2023} as student, which is an efficient version of ViT designed for mobile devices. All our student models are pre-trained on Tiny ImageNet, a subset of ImageNet that was introduced by \cite{le-tiny-2015}.

\subsubsection{Baselines} 
We compare our approach with four state-of-the-art model stealing methods \citep{barbalau-NeurIPS-2020,orekondy-cvpr-2019,wang-ICML-2021, zhang-ICLR-2022}. The first baseline is Black-Box Ripper \citep{barbalau-NeurIPS-2020}, a framework that employs a generative model in order to create proxy data, but the framework is rather focused on achieving a high accuracy, irrespective of the number of API calls. The second baseline is IDEAL \citep{zhang-ICLR-2022}, a method that also generates synthetic data to train the student. However, IDEAL rather aims to reduce the number of API calls to the teacher.

Knockoff Nets \citep{orekondy-cvpr-2019} represent our third baseline. Aside from their relevance in the model stealing research, Knockoff Nets have a similar focus to our own, namely to optimize the number of teacher (or victim) passes. Following \cite{orekondy-cvpr-2019}, we use CIFAR-100 as proxy data for Knockoff Nets, when the evaluation is performed on CIFAR-10. Similarly, we use ImageNet-200 \citep{Russakovsky-IJCV-2015} (a subset of 200 classes from ImageNet) as proxy data for Food-101 and FER+.

The final baseline is called Zero-Shot Decision-Based Black-Box Knowledge Distillation (ZSDB3KD) \citep{wang-ICML-2021}, which focuses on performance. Although the authors do not impose any limits on the number of calls, they treat the teacher as a black-box model and present a method that does not utilize the original data set, hence being a zero-shot method.

For a fair comparison, we impose the same limit on the number of API calls for all frameworks. Moreover, we use the same teacher and student architectures for all frameworks. For all baselines, we employ student models pre-trained on Tiny ImageNet, for a fair comparison with our method. Hence, the reported accuracy rates reflect the performance levels of the training frameworks, namely Black-Box Ripper \citep{barbalau-NeurIPS-2020}, Knockoff Nets \citep{orekondy-cvpr-2019}, IDEAL \citep{zhang-ICLR-2022}, ZSDB3KD \citep{wang-ICML-2021} and ASPKD (ours).


\subsubsection{Hyperparameters} 
Throughout the experiments, we employ the Adam optimizer \citep{Kingma-ICLR-2015} with a decaying learning rate scheduler. The hyperparameters for the teachers are tuned independently of the students, thus preserving the black-box nature of the teachers. In the case of CIFAR-10, the teachers are trained for $100$ epochs with early stopping and a learning rate of $5\cdot 10^{-4}$ on mini-batches of $64$ samples, while the scheduler has a step size of $5$ with $\gamma=0.95$. For the experiments on Food-101, the teachers are trained for $100$ epochs with a learning rate of $10^{-3}$ and a mini-batch size of $64$. For the learning rate scheduler, the step size is $20$ with $\gamma=0.95$. For the experiments on FER+, the teachers are trained for $30$ epochs with a learning rate of $5 \cdot 10^{-4}$ and a mini-batch size of $32$. For the learning rate scheduler, the step size is $10$ with $\gamma=0.55$, employing a momentum of $0.2$.

The student models are fine-tuned on $15\%$ of the proxy data. The students are trained for $100$ epochs with early stopping on mini-batches of $64$ samples. As far as the active learning strategy is concerned, we set the value of $\sigma$ in Eq.~\eqref{eq:dist2prob} to $17$. The nearest neighbors algorithm in the self-paced learning method uses $k=5$ neighbors. The optimal values for the other hyperparameters of the students on all data sets are reported in Table~\ref{tab:hyperparameters}. 

We present results with two versions for our framework: one that learns from hard teacher labels, and one that learns from soft teacher labels. For ASPKD based on hard labels, we use the Euclidean distance to find the nearest neighbors during self-paced learning. For ASPKD based on soft labels, we use the cosine distance defined in Eq.~\eqref{eq_cosdist}.

\begin{table*}[t]
\centering
\caption{Optimal hyperparameters of the student models for all three data sets.}
\label{tab:hyperparameters}
\begin{tabular}{llcccc}
    \toprule
    \multirow{2}{*}{Data set} & \multirow{2}{*}{Student} & \multirow{2}{*}{Diffusion model} & Learning & Step & \multirow{2}{*}{$\gamma$} \\
        & &  & rate & size & \\
    \midrule
   \multirow{4}{*}{CIFAR-10} & Half-AlexNet & GLIDE & $9\cdot 10^{-4}$ & $20$ & $0.95$ \\
    & ResNet-18 & GLIDE & $9\cdot 10^{-4}$ & $20$ & $0.95$ \\
    & Half-AlexNet & Stable Diffusion & $6\cdot 10^{-4}$ & $20$ & $0.95$ \\
    & ResNet-18 & Stable Diffusion & $10^{-4}$ & $30$ & $0.9$ \\
    \midrule
   \multirow{2}{*}{Food-101} & Half-AlexNet & Stable Diffusion & $7\cdot 10^{-5}$ & $10$ & $0.95$ \\
    & ResNet-18 & Stable Diffusion & $10^{-4}$ & $30$ & $0.95$ \\
    \midrule
    FER+ & FastViT & SDXL & $10^{-4}$ & $10$ & $0.5$ \\
  \bottomrule
\end{tabular}
\end{table*}






\subsubsection{Evaluation}

All models are evaluated on the official test sets of CIFAR-10, Food-101, and FER+. For the teacher models, we report the classification accuracy with respect to the ground-truth labels. Since the goal of the student models is to replicate the teachers, we evaluate each student in terms of the classification accuracy with respect to the labels predicted by its teacher. For each experiment, we report the average performance computed over 5 runs with each model. 


\subsection{Main Results} 

\begin{figure*}[t]
\begin{center}
\centerline{\includegraphics[width=1.0\linewidth]{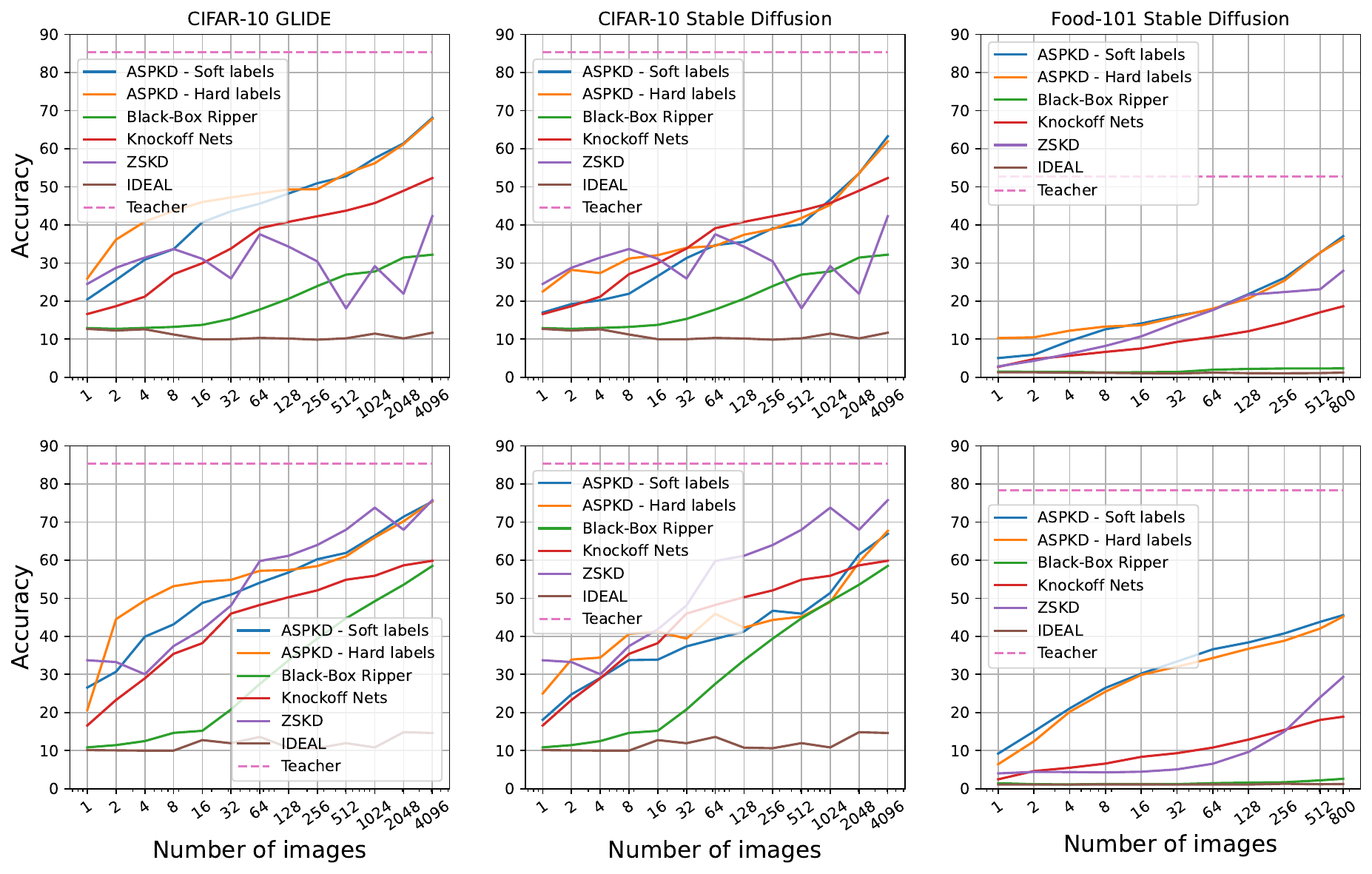}}
\caption{Empirical results for six experimental scenarios, where the maximum number of API calls per class takes values in the set $\{1,2,4,...,4096\}$ for CIFAR-10, and the set $\{1,2,4,...,512,800\}$ for Food-101. The plots on the first two columns depict the results on CIFAR-10 \citep{Krizhevsky-TECHREP-2009}, while the plots on the last column illustrate the results on Food-101 \citep{Bossard-ECCV-2014}. The proxy data used by ASPKD for the plots on the first column is generated with GLIDE \citep{Nichol-ICML-2021b}, while the proxy data used by ASPKD for the other plots is generated with Stable Diffusion v2 \citep{Rombach-CVPR-2022}. For the plots on the top row, the student architecture is based on Half-AlexNet. For the plots on the bottom row, the student model is ResNet-18. We compare the results of ASPKD based on soft and hard labels with those of four state-of-the-art frameworks: Black-Box Ripper \citep{barbalau-NeurIPS-2020}, Knockoff Nets \citep{orekondy-cvpr-2019}, IDEAL \citep{zhang-ICLR-2022} and ZSDB3KD \citep{wang-ICML-2021}. For reference, the accuracy rate of the corresponding teacher model is added to each plot. For each experiment, we report the average performance computed over 5 runs with each model. Best viewed in color.}
\label{all_results}
\end{center}
\end{figure*}

\begin{figure*}[t]
\begin{center}
\centerline{\includegraphics[width=0.60\linewidth]{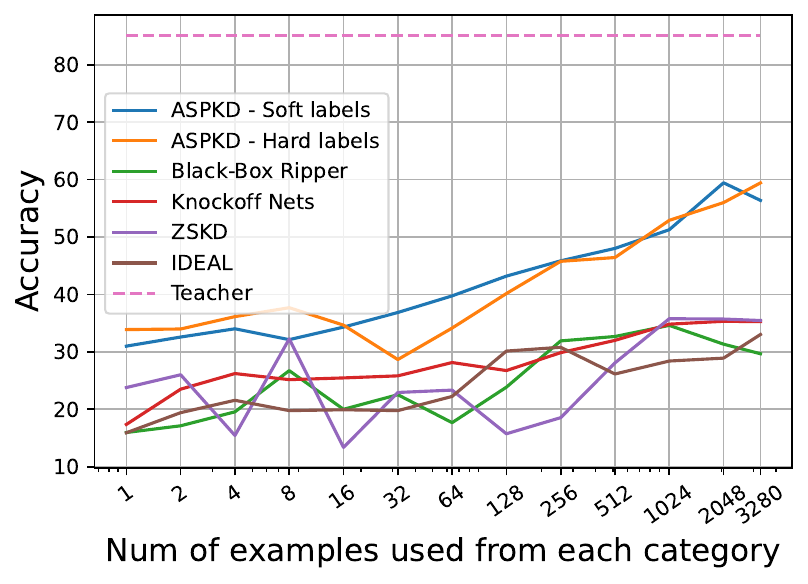}}
\caption{The experimental results for the FER+ data set, employing ViT-B as teacher and FastViT as student. We compare the results of ASPKD based on soft and hard labels with those of four state-of-the-art frameworks: Black-Box Ripper \citep{barbalau-NeurIPS-2020}, Knockoff Nets \citep{orekondy-cvpr-2019}, IDEAL \citep{zhang-ICLR-2022} and ZSDB3KD \citep{wang-ICML-2021}. For reference, we add the accuracy rate of the ViT-B teacher model. For each number of API calls, we report the average performance computed over 5 runs with each model. Best viewed in color.}
\label{vit_results}
\end{center}
\end{figure*}

On CIFAR-10, we have four evaluation scenarios, since there are two teacher-student pairs and two diffusion models. On Food-101, we have two evaluation scenarios, as we use the same teacher-student pairs, but only one diffusion model. Finally, on FER+, we consider one scenario which involves transformer-based models. In total, we compare our framework (ASPKD) with Black-Box Ripper \citep{barbalau-NeurIPS-2020}, Knockoff Nets \citep{orekondy-cvpr-2019}, IDEAL \citep{zhang-ICLR-2022} and ZSDB3KD \citep{wang-ICML-2021} in seven scenarios. The maximum number of API calls per class takes values in the set $\{1,2,4,...,4096\}$ for the four CIFAR-10 scenarios, the set $\{1,2,4,...,512,800\}$ for the two Food-101 scenarios, and the set $\{1,2,4,...,2048,3280\}$ for the FER+ scenario. The corresponding results are presented in Figures \ref{all_results} and \ref{vit_results}. The synthetic training data is generated using GLIDE for the plots on the first column in Figure \ref{all_results}, and Stable Diffusion for the rest. For the plots on the first row in Figure \ref{all_results}, the teacher-student pair is represented by AlexNet$\rightarrow$Half-AlexNet. For the second row, the teacher-student pair is ResNet-50$\rightarrow$ResNet-18. In each plot, we present results with two versions for ASPKD, corresponding to the type of labels returned by the teacher, soft or hard. 

When compared to Black-Box Ripper \citep{barbalau-NeurIPS-2020}, our framework (ASPKD) obtains significantly better results in all seven evaluation scenarios, regardless of the maximum number of API calls. In five evaluation scenarios (illustrated on the first and third columns in Figure \ref{all_results}, as well as in Figure \ref{vit_results}), both ASPKD versions outperform Knockoff Nets \citep{orekondy-cvpr-2019} by considerable margins. For the other two scenarios, where the proxy data for CIFAR-10 is generated with Stable Diffusion, Knockoff Nets \citep{orekondy-cvpr-2019} temporarily surpass ASPKD, when the number of API calls per class ranges between 32 and 1024. ASPKD based on hard labels yields better results than Knockoff Nets in the more challenging few-call settings, namely when the maximum number of API calls per class is below 16. As the number of API calls per class increases, the two ASPKD versions register faster performance gains, recovering the temporary performance gap and even outperforming Knockoff Nets when the number of API calls per class is at least 2048. Furthermore, it can be observed that IDEAL \citep{zhang-ICLR-2022} is consistently performing worse than ASPKD. The main justification behind this observation being that its generator is not producing diverse samples under the restricted the number of API calls, which limits the number of training iterations of the generator. Our method also surpasses ZSDB3KD \citep{wang-ICML-2021} in most cases, except a few evaluation scenarios on CIFAR-10, when ResNet-18 is used as the student. Nevertheless, on the more challenging Food-101 (see last column in Figure \ref{all_results}) and FER+ (see Figure \ref{vit_results}) data sets, ZSDB3KD is consistently below our method. When applied to vision transformers, our approach significantly outperforms its competitors (see Figure \ref{vit_results}). 

Considering the bigger picture, we conclude that ASPKD leads to generally better results, surpassing Black-Box Ripper \citep{barbalau-NeurIPS-2020}, Knockoff Nets \citep{orekondy-cvpr-2019}, IDEAL \citep{zhang-ICLR-2022} and ZSDB3KD \citep{wang-ICML-2021} in most evaluation scenarios.

\subsection{Ablation Studies}
\begin{figure*}[t]
\begin{center}
\centerline{\includegraphics[width=1.0\linewidth]{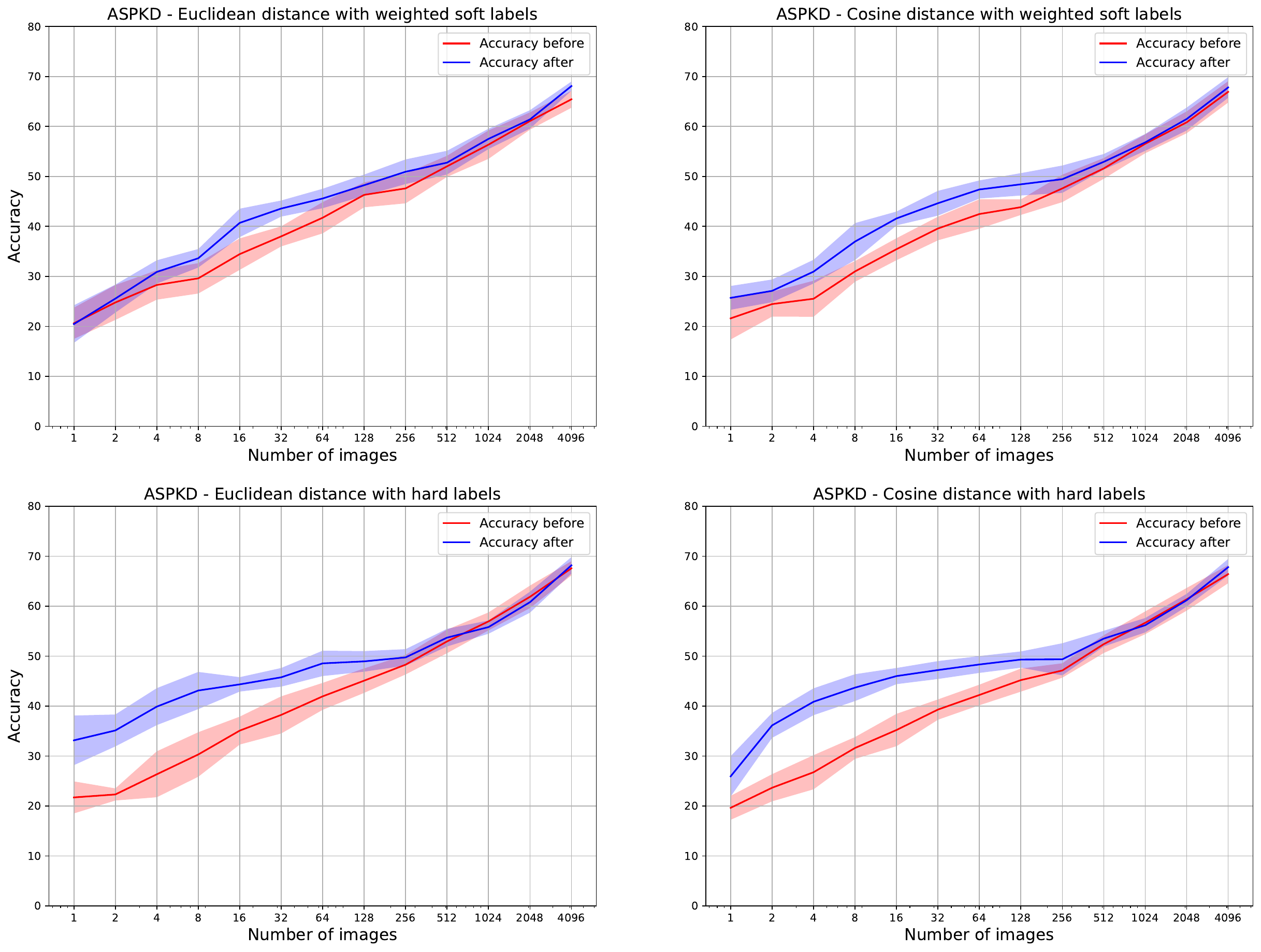}}
\caption{Accuracy rates before and after introducing our self-paced learning strategy. The comparison is carried out for four combinations of distance functions (Euclidean or cosine) and label types (hard or soft). For each experiment, we present the mean accuracy and the standard deviation over 5 runs. The test data is CIFAR-10, and the synthetic training images are generated by GLIDE \citep{Nichol-ICML-2021b}. The teacher is AlexNet, and the student is Half-AlexNet. Best viewed in color.}
\label{before_after}
\end{center}
\end{figure*}

In order to demonstrate the capability of the individual stages of the proposed method, we carried out several ablation studies. To demonstrate the efficiency of our self-paced learning scheme, we conduct an analysis of the performance before and after introducing our self-paced strategy, considering the four possible combinations of distance functions (Euclidean or cosine) and teacher labels (hard or soft). The corresponding results are illustrated in Figure~\ref{before_after}. A clear pattern emerges when analyzing the four plots, specifically that self-paced learning brings considerable performance gains when the number of API calls per class is below 512. The improvements are usually higher for ASPKD based on hard labels. In conclusion, the empirical evidence indicates that our self-paced learning strategy plays a key role in the few-call model stealing scenarios.

In Table~\ref{tab:acc_each_component}, we present the performance impact caused by alternatively and jointly introducing the active learning and the self-paced learning strategies, respectively. When we separately introduce the active learning and the self-paced learning strategies, we observe that each strategy brings significant gains in the majority of cases. Interestingly, when the number of API calls per class is below or equal to 128, we notice even higher gains when both strategies are jointly introduced. In summary, the results show the benefits of both strategies.

\begin{table*}[th]
  \centering
  \caption{Accuracy rates of the Half-AlexNet student based on various training procedures. The vanilla procedure is based on training the student on proxy data with teacher labels in a conventional way. Next, we show the impact of separately and jointly introducing active learning and self-paced learning (based on cosine distance and soft labels), respectively. For each experiment, we present the mean accuracy and the standard deviation over 5 runs. The results are reported on CIFAR-10, while the proxy training data is generated by GLIDE \citep{Nichol-ICML-2021b}.}
  \label{tab:acc_each_component}
  \begin{tabular}{ccccc}
    \toprule
    $\;\;$\#Samples$\;\;$ &  & \multirow{2}{*}{$\;\;\;$+ Active$\;\;\;$} & \multirow{2}{*}{+ Self-paced} & $\;\;$+ Active \&$\;\;$\\
        per  & $\;\;\;$Vanilla$\;\;\;$ & \multirow{2}{*}{learning} & \multirow{2}{*}{learning} & self-paced\\
         class   &  &  & & learning \\
    \midrule
     1    & 19.4$\pm$2.1 & 20.7$\pm$3.6 & 25.0$\pm$2.7 & 25.9$\pm$4.0\\
     2    & 25.9$\pm$3.7 & 26.5$\pm$2.0 & 25.6$\pm$1.1 & 36.1$\pm$2.5\\
     4    & 27.7$\pm$2.6 & 29.9$\pm$1.2 & 28.5$\pm$0.5 & 40.9$\pm$2.7\\
     8    & 29.4$\pm$2.0 & 33.0$\pm$2.0 & 35.3$\pm$1.7 & 43.7$\pm$2.7\\
     16   & 36.3$\pm$3.5 & 41.8$\pm$3.0 & 40.8$\pm$1.1 & 46.0$\pm$1.6\\
     32   & 39.2$\pm$2.1 & 43.9$\pm$1.3 & 46.4$\pm$1.7 & 47.2$\pm$1.8\\
     64   & 43.6$\pm$1.4 & 46.0$\pm$2.7 & 45.8$\pm$1.2 & 48.3$\pm$1.7\\
     128  & 46.6$\pm$1.8 & 48.8$\pm$1.5 & 48.0$\pm$1.6 & 49.3$\pm$1.6\\
     256  & 47.7$\pm$3.2 & 50.3$\pm$1.9 & 50.0$\pm$1.5 & 49.4$\pm$3.2\\
     512  & 52.5$\pm$2.1 & 53.6$\pm$1.9 & 53.6$\pm$0.8 & 53.5$\pm$1,6\\
     1024 & 55.5$\pm$2.5 & 56.0$\pm$1.5 & 56.6$\pm$2.2 & 56.2$\pm$1.4\\
     2048 & 60.3$\pm$1.7 & 61.5$\pm$2.1 & 63.1$\pm$0.9 & 61.2$\pm$1.3\\
     4096 & 66.3$\pm$1.8 & 68.0$\pm$0.7 & 67.7$\pm$1.3 & 67.8$\pm$1.7\\
  \bottomrule
\end{tabular}
\end{table*}

\begin{figure*}[htbp]
\begin{center}
\centerline{\includegraphics[width=0.7\linewidth]{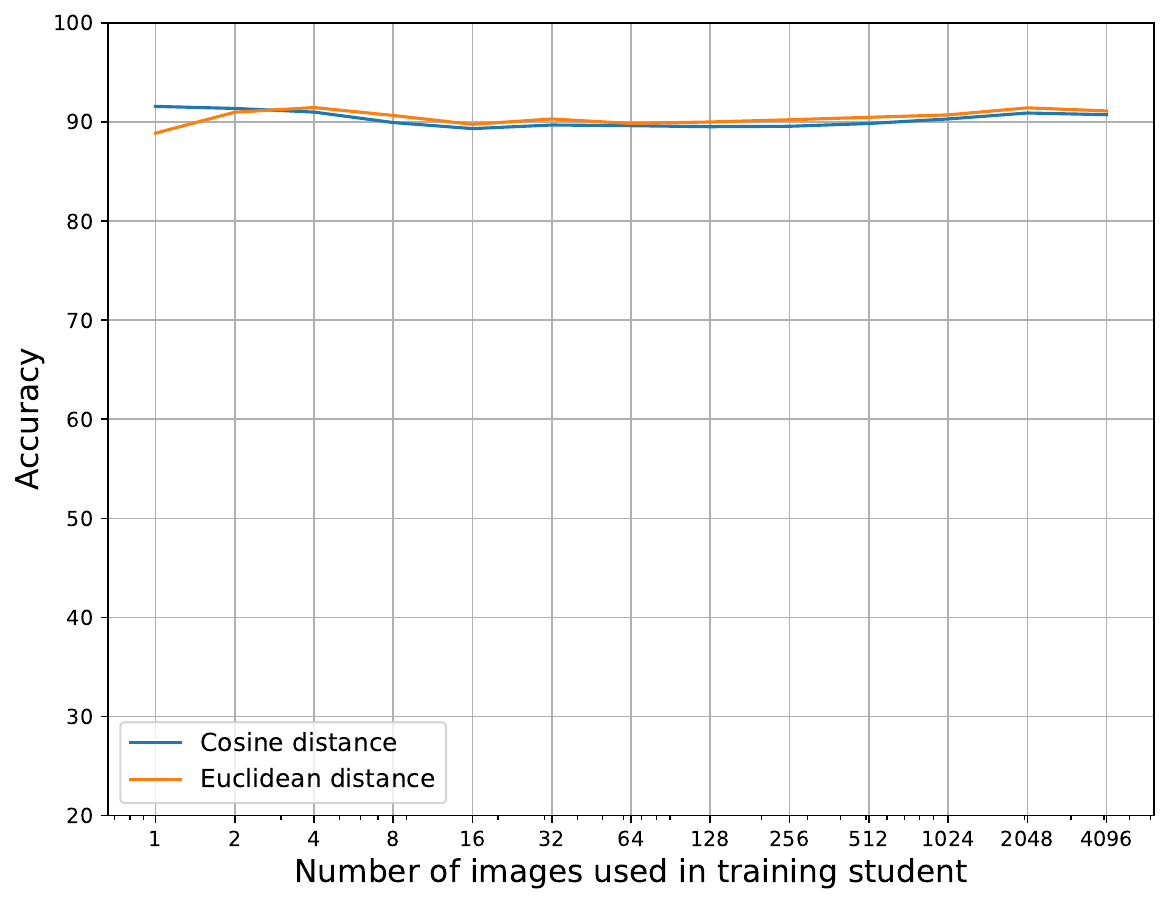}}
\caption{The accuracy of the pseudo-labels assigned during the self-paced learning process, with respect to the labels that would have been predicted by the teacher. The results are reported for the Half-AlexNet student on CIFAR-10, while the proxy training data is generated by GLIDE \citep{Nichol-ICML-2021b}.}
\label{self_paced_acc}
\end{center}
\end{figure*}

One of the dangers of self-paced learning is degrading the performance because of the amount of noise in the pseudo-labels. Given the uncertainty of the pseudo-labeling process, we next analyze the accuracy of the pseudo-labels with respect to the labels that would have been predicted by the teacher model. The corresponding results, which are shown in Figure~\ref{self_paced_acc}, indicate that the quality of the pseudo-labels is consistently high, regardless of the number of API calls per class. This explains why our self-paced learning strategy works so well.


Another interesting experiment is to only utilize the labels with which the images were created (the class names included in the prompts), thus eliminating the need to call the black-box model. We showcase this training scenario in Table~\ref{tab:original_labels}, using the ResNet-18 student on the CIFAR-10 data set, with the proxy data generated by GLIDE. The accuracy rates obtained by the conventional training method (based on the original labels) are significantly worse than those obtained by ASPKD. This clearly shows the necessity to make calls to the black-box model in order to replicate its functionality. Furthermore, in the same table, we report the execution time of both methods, which allows us to estimate the additional training time required by our method. At a first glance, it appears that our method can require up to $9\times$ more training time, where the larger time differences are observed when the number of API calls is typically lower. The higher differences observed with lower numbers of samples are due to the resulting high numbers of synthesized data points for which pseudo-labels are assigned during self-paced learning. However, if we consider the training time required to reach a certain accuracy level, our method can actually be more efficient. For example, the conventional regime reaches an accuracy of $52.68\%$ with 4096 samples per class after about 5592 seconds of training, while ASPKD reaches an accuracy of $54.08\%$ with 64 samples per class after about 768 seconds of training. Hence, in this case, our method is $7\times$ faster than conventional training.

\begin{table*}[th]
\centering
\setlength\tabcolsep{0.20cm}
\caption{Comparing performance (in percentages) and time (in seconds) between our method and a conventional training pipeline that uses the original labels with which the artificial data was created (involving no calls to the teacher). The experiments are carried out on CIFAR-10, with the ResNet-18 student trained on proxy data generated by GLIDE. The training time is measured on an NVIDIA GeForce GTX 3090 GPU with 24 GB of VRAM. For each experiment, we present the mean accuracy and the mean time over 5 runs. The corresponding standard deviations are also reported.}
\label{tab:original_labels}
\begin{tabular}{ccccc}
    \toprule
    {\#Samples } & \multicolumn{2}{c}{Accuracy} & \multicolumn{2}{c}{Time} \\
       per class & Conventional  & ASPKD & Conventional  & ASPKD \\
    \midrule
     1    & $7.72\pm0.12$  & $26.56\pm3.91$ & $70.51\pm45.53$     & $583.94\pm104.80$ \\
     2    & $8.48\pm0.12$  & $30.71\pm3.54$ & $72.97\pm45.89$     & $616.39\pm144.92$ \\
     4    & $7.97\pm0.09$  & $39.90\pm4.83$ & $77.30\pm43.25$      & $595.55\pm92.75$ \\
     8    & $8.10\pm0.05$   & $43.14\pm2.98$ & $81.29\pm44.54$     & $635.14\pm136.70$ \\
     16   & $7.98\pm0.14$  & $48.77\pm1.93$ & $97.19\pm50.31$     & $670.44\pm153.55$ \\
     32   & $9.58\pm0.42$  & $50.95\pm2.20$ & $119.33\pm53.45$    & $722.84\pm169.09$ \\
     64   & $10.65\pm0.14$ & $54.08\pm2.58$ & $165.94\pm69.40$     & $768.08\pm131.92$ \\
     128  & $15.60\pm0.49$  & $56.77\pm2.27$ & $247.29\pm105.85$   & $837.59\pm144.93$ \\
     256  & $25.74\pm0.67$ & $60.24\pm0.93$ & $414.46\pm176.00$    & $985.83\pm283.60$ \\
     512  & $37.52\pm0.08$ & $61.90\pm1.05$ & $764.72\pm345.00$    & $1386.35\pm540.56$ \\
     1024 & $46.63\pm0.15$ & $66.45\pm1.18$ & $1452.19\pm644.55$  & $2616.69\pm1123.48$ \\
     2048 & $51.38\pm2.35$ & $71.39\pm0.79$ & $2732.74\pm1328.19$ & $4797.63\pm882.25$ \\
     4096 & $52.68\pm0.15$ & $75.37\pm0.32$ & $5591.87\pm2791.61$ & $5922.99\pm2406.20$ \\
  \bottomrule
\end{tabular}
\end{table*}


\section{Ethical Considerations}

The possibility of launching model stealing attacks against machine learning models exposed via public APIs is a serious threat for companies releasing such models. Our results show that there is a high risk of stealing the intellectual property behind the released models, even when the access is restricted to just the output of the respective models. Indeed, we demonstrate how accessible it is to perform a model stealing attack, relying only on public information and accessible resources. We used three different open-source diffusion models to generate proxy data. Then, by querying the black-box model and obtaining its hard or soft labels for a limited number of images, our method can easily be applied to distill the knowledge of the black-box teacher into a copy model, which can later be used with no restrictions. This type of attack can be launched by any machine learning engineer, resulting in a potentially large number of intellectual property infringements. We consider that our work will inspire current researchers to continue on this track and work towards discovering methods of prevention against model stealing attacks. In this way, companies and individuals that publicly release models will benefit from enhanced security mechanisms.

\section{Conclusion}

In this study, we explored the task of replicating the functionality of black-box machine learning models. We designed our method to be applicable in real-world scenarios, where there are several constraints, \ie~no access to the training set, no information about the architecture of the victim model or about its training process, as well as a cap on the number of permitted model calls. Our first contribution was to generate synthetic training data using a text-to-image diffusion model, allowing us to generate any class entity, while having a high diversity of images. Due to the limit imposed on the number of API calls, we introduced a self-paced learning method that assigns pseudo-labels for generated images that never get passed through the black-box teacher model. We also presented an active learning strategy that improves the process of selecting the proxy data to be labeled by the teacher. We carried out extensive experiments focusing on reducing the number of API calls, reporting results on various test cases based on multiple combinations of teacher-student architectures, distinct data sets, different diffusion models, and different output types given by the black-box model. In the current surge of artificial intelligence solutions, our research aims to raise awareness of the exposure to model stealing attacks.

In future work, we aim to investigate methods to prevent few-call model stealing attacks.

\backmatter

\section*{Declarations}

\textbf{Funding} The authors did not receive support from any organization for the submitted work.

\vspace{0.5cm}
\noindent
\textbf{Conflict of interest} The authors have no conflicts of interest to declare that are relevant to the content of this article.

\vspace{0.5cm}
\noindent
\textbf{Ethics approval} Not applicable.

\vspace{0.5cm}
\noindent
\textbf{Consent to participate} Not applicable.

\vspace{0.5cm}
\noindent
\textbf{Consent for publication} The authors give their consent for publication.

\vspace{0.5cm}
\noindent
\textbf{Authors' contributions} The authors have contributed equally to the work.

\vspace{0.5cm}
\noindent
\textbf{Availability of data and materials} The data sets are publicly available online.

\vspace{0.5cm}
\noindent
\textbf{Code availability} The code has been made publicly available for non-commercial use at \url{https://github.com/vladhondru25/model-stealing}.

\vspace{0.5cm}
\noindent
\textbf{Open Access} This article is licensed under a Creative Commons Attribution 4.0 International License, which permits use, sharing, adaptation, distribution and reproduction in any medium or format, as long as you give appropriate credit to the original author(s) and the source, provide a link to the Creative Commons licence, and indicate if changes were made. The images or other third party material in this article are included in the article’s Creative Commons licence, unless indicated otherwise in a credit line to the material. If material is not included in the article’s Creative Commons licence and your intended use is not permitted by statutory regulation or exceeds the permitted use, you will need to obtain permission directly from the copyright holder. To view a copy of this licence, visit \url{http://creativecommons.org/licenses/by/4.0/}.


\bibliography{sn-bibliography}

\end{document}